\documentclass[11pt]{article}

\usepackage{ranlp97}
\usepackage{epsf}

\title{Combining Multiple Methods for the Automatic Construction of Multilingual WordNets\thanks {This
research has been partially funded by the Spanish Research Department
(ITEM project TIC96-1243-C03-03), the Catalan Research department (CIRIT 1995SGR 00566) and EU Comission (EuroWordNet LE4003)}.}

\author{Jordi Atserias, Salvador Climent, Xavier Farreres,
	German Rigau, Horacio Rodr\'{\i}guez\\
        Departament de Llenguatges i Sistemes Informatics\\
	Universitat Politecnica de Catalunya.\\
        Carrer Jordi Girona Salgado, 1-3. 08034 Barcelona, Catalonia \\
	\texttt{\{batalla,farreres,g.rigau,horacio\}@lsi.upc.es, climent@lingua.fil.ub.es}
	}

\begin{document}

\maketitle

\begin{abstract}
This paper explores the automatic construction of a multilingual Lexical Knowledge Base from preexisting lexical resources. First, a set of automatic and complementary techniques for linking Spanish words collected from monolingual and bilingual MRDs to English WordNet synsets are described. Second, we show how resulting data provided by each method is then combined to produce a preliminary version of a Spanish WordNet with an accuracy over 85\%. The application of these combinations results on an increment of the extracted connexions of a 40\% without losing accuracy. Both coarse-grained (class level) and fine-grained (synset assignment level) confidence ratios are used and evaluated. Finally, the results for the whole process are presented.
\end{abstract}

\section{Introduction}

There is no doubt about the increasing importance of using wide coverage ontologies for NLP tasks. Although available ontologies (Upper Model \cite{Bateman'90}, CYC \cite{Lenat'95}, WordNet \cite{Miller'90}, ONTOS \cite{Nirenburg+Defrise'93}, Mikrokosmos, EDR \cite{Yokoi'95}, etc.)\footnote{ See an overview and discussion of CYC, WordNet and EDR systems in Communications of the ACM 38:(11), pages 33-48, 1995.} differ in great extent on several characteristics  (e.g. broad coverage vs. domain specific, lexically oriented vs. conceptually-oriented, granularity, kind of information placed in nodes, kind of relations, way of building, etc.), it is clear that WordNet has become a de-facto standard for a wide range of NL applications. Developed at Princeton by George Miller and his research group \cite{Miller'90}, the figures the currently available version of WordNet 1.5 (WN1.5) shows are impressive (119,217 words, 91,587 synsets). WN1.5 is organised as a network of lexicalized concepts (Synsets) which are sets of word meanings (WMs) considered to be synonymous within a context. Synsets are connected by several semantic relations (nevertheless, only that of hypernymy-hyponymy is considered in this work).

WordNet success has encouraged several projects in order to build WordNets (WNs) for other languages or to develop multilingual WNs. The most ambitious of such efforts is EuroWordNet (EWN)\footnote{EuroWordNet: Project LE- 4003 of the EU.}, a project aiming to build a multilingual WordNet for several European languages\footnote{Initially three languages, apart from English, were involved: Dutch, Italian and Spanish. The project has been recently extended for covering French and German.}. The work we present here is included within EWN and presents our approach for (semi)automatically building a Spanish WN \cite{Climent+'96}. The main strategy within our aproach is to map Spanish words to WN1.5 synsets, creating for Spanish a parallel-in-structure network. Therefore, our main goal is to attach Spanish word meanings to the existing WN1.5 concepts. This paper describes automatic techniques which  have been developed in order to achieve this goal for nouns.

Recently, several attemps have been performed to produce automatically multilingual ontologies. \cite{Ageno+'94} link taxonomic structures derived from DGILE and LDOCE by means of a bilingual dictionary. \cite{Knight+Luk'94} focus on the construction of Sensus, a large knowledge base for supporting the Pangloss Machine Translation system, merging ontologies (ONTOS and UpperModel) and WordNet with monolingual and bilingual dictionaries. \cite{Okumura+Hovy'94} describe a (semi)automatic method for associating a Japanese lexicon to an ontology using a Japanese/English bilingual dictionary as a "bridge". \cite{Rigau+'95} link Spanish word senses to WordNet synsets using also a bilingual dictionary. \cite{Rigau+Agirre'95} exploit several bilingual dictionaries for linking Spanish and French words to WordNet synsets.

Our approach for building the Spanish WN (SpWN) is based on the following considerations:

\begin{itemize}

\item The close conceptual similarity of English and Spanish allows for the preservation of the structure of WN1.5 in order to build the SpWN. Moreover, when necessary, lexicalization mismatches are solved using multi-word traslations (collocations) supplied by bilingual dictionaries.
\item An extensive use of pre-existing structured lexical sources is performed in order to achieve a massive automatic acquisition process.
\item The accuracy of cross-language mappings is validated by hand on a sample. Each attachment to WN bears a confidence score (CS).
\item Only attachments over a threshold are considered. Moreover, a manual inspection of attachments in a given range will be carried out.
\end{itemize}

Undoubtfully, following this aproach most of the criticisms placed to WN1.5 also apply to SpWN: too much sense fine-grainedness, lack of cross-POS relationships, simplicity of the relational information, not purely lexical but rather lexical-conceptual database, etc. Despite of these drawbacks, WN1.5 is widely used and tested and supports few but the most basic semantic relations. Our aproach ensures that most of the huge networking effort, which is necessary to build a WN from scratch, is already done.

The different sources involved in the process show a different accuracy. High CSs can be assigned to original sources, as MRDs, but derived sources, which result  from the performance of automatic procedures, come to bear substantially lower CSs. Our major claim is that multiple source/procedures leading to the same result will increase the particular CS while when leading to different results the overall CS will  decrease.

This paper is organized as follows. In section 2 Lexical Knowledge resources used are presented. Section 3 describes the different types of extraction/mapping methods developed. Main results and quality assesments issues are presented in Section 4. Section 5 presents some final remarks.

\section{Lexical Knowledge Sources}

\begin{table*}[htb]
\centering
\begin{tabular}{l|rrrr}
% \hline
\textsc{} & \textsc{English nouns} & \textsc{Spanish nouns} & \textsc{Synsets} & \textsc{Connections}\\
\hline
WordNet1.5 			& 87,642 & -	 	& 60,557	& 107,424\\
Spanish/English			& 11,467 & 12,370	& -		& 19,443\\
English/Spanish			& 10,739 & 10,549	& -		& 16,324\\
Hbil				& 15,848 & 14,880	& -		& 28,131\\
Maximum Reacheable Coverage	& 12,665 & 13,208	& 19,383	& 66,258\\
of WordNet			& 14\%	 & -		& 32\%		& -\\
of bilingual			& 80\%	 & 90\%		& -		& -\\
% \hline			
\end{tabular}
\caption{\label{dictionarystatisctics}: Dictionary Statistics}
\label{dictionarystatistics:table:blah-blah}
\end{table*}

Several lexical sources have been applied in order to assign Spanish WMs to WN1.5 synsets:
\begin{enumerate}
\item Spanish/English and English/Spanish bilinguals\footnote{Diccionario Vox/Harraps Esencial Espa\~nol/Ingl\'es - Ingl\'es/Espa\~nol  Biblograf S.A. Barcelona 1992}
\item A large Spanish monolingual dictionary\footnote{DGILE: Diccionario General Ilustrado de la Lengua Espa\~nola - Vox - M.Alvar (ed) Biblograf. S.A.  Barcelona 1987}
\item English WordNet (WN1.5).
\end{enumerate}

By merging both directions of the bilingual dictionaries what we call homogeneous bilingual (HBil) has been obtained. The maximum synset coverage we can expect to reach by using HBil due to its small size is 32\%. In table \ref{dictionarystatisctics}\footnote{Connections can be word/word or word/synset. When there are synsets involved the connections are Spanish-word/synset,(except for WordNet itself), otherwise Spanish-word/English-word.} the summarised amount of data is shown.

\section{Methods}

Bilingual entries must be disambiguated against WN. The different procedures developed for linking Spanish lexical entries to WN synsets can be classified in three main groups according to the kind of knowledge sources involved in the process:

\begin{itemize}
\item Class methods: use as knowledge sources individual entries coming from bilinguals and WN synsets.
\item Structural methods: take profit of the WN structure.
\item Conceptual Distance methods: makes use of knowledge relative to meaning closeness between lexical concepts.
\end{itemize}

Every method has been manually inspected in order to measure its CS. Such tests have been performed on a random sample of 10\% using the Validation Interface (VI), an enviroment designed to allow hand validation of Spanish word forms to WN synsets assignment. It allows to consult and to navigate through the monolingual and bilingual lexical databases and WN. The following diagnostics can result from the performance of  this validation:

\begin{description}
\item[ok]: correct links.
\item[ko]: fully incorrect links.
\item[hypo]: links to a hyponym of the correct synset.
\item[hyper]: links to a hyperonym of the correct synset.
\item[near]: links to near synonyms that could be considered ok.
\end{description}

\subsection{Class Methods}

Following the properties described in \cite{Rigau+Agirre'95} Hbil has been processed and 2 groups of 4 different cases have been collected depending on whether the English words are either monosemous or polysemous  relative to WN 1.5. Afterwards two hybrid criteria are considered as well.

\subsubsection{Monosemic Criteria}
These criteria apply only to monosemous EW with respect to WN1.5. As a result,  this unique synset is linked to the corresponding Spanish words.
\begin{itemize}

\item Monosemic-1 criterion (1:1) :
\begin{figure}[htb]
\hfil\epsfbox{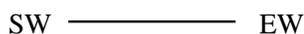}\hfil
\caption{Monosemic Criteria}
\end{figure}

A Spanish Word (SW) has only one English translation (EW); symmetrically, EW has SW as its unique traslation.

\item Monosemic-2 criterion (1:N with N$>$1):
\begin{figure}[htb]
\hfil\epsfbox{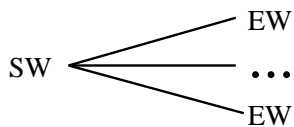}\hfil
\caption{Monosemic-2 Criteria}
\end{figure}

A SW has more than one translation; each EW has SW  as its unique traslation.

\item Monosemic-3 criterion (M:1 with M$>$1):
\begin{figure}[htb]
\hfil\epsfbox{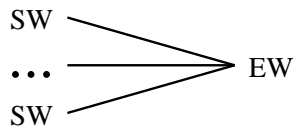}\hfil
\caption{Monosemic-3 Criteria}
\end{figure}

Several SWs have the same translation; EW has several translations to Spanish.

\item Monosemic-4 criterion (M:N with M$>$1 \& N$>$1):
\begin{figure}[htb]
\hfil\epsfbox{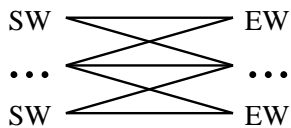}\hfil
\caption{\label{mono4} Monosemic-4 Criteria}
\label{mono4:figure:bla}
\end{figure}

Several SWs have different translations; EWs also have several translations.

\end{itemize}
\noindent

\subsubsection{Polysemic Criteria}

These criteria follow the four criteria descrived in previous subsection but for polysemous English words (relative to WN1.5).

\subsubsection{Hybrid Criteria}

\begin{itemize}

\item Variant criterion

For a WN1.5 synset which contains a set of variants EWs, if it is the case that two or more of the variants EWi  have only one translation to the same Spanish word SW, a link is produced for SW into the WN1.5 synset.

\item Field criterion

This procedure makes use of the existence of a field identifier in
some entries (over 4,000) of the English/Spanish bilingual. For each
English entry bearing a field identifier (EW), if it is the case that
both occur in the same synset, for each EW translation to Spanish a
link is produced. Results of the manual verification for each
criterion are shown in table \ref{manualverification}. 

\begin{table*}[htb]
\centering
\begin{tabular}{l|rrrrrrrr}
% \hline
\textsc{Criterion} & \textsc{\#Links} & \textsc{\#Synsets} & \textsc{\#Words} & \textsc{\%ok} & \textsc{\%ko} & \textsc{\%hypo} & \textsc{\%hyper} & \textsc{\%near}\\
\hline
mono1 & 3,697 & 3,583 & 3,697 & 92 & 2 & 2 & 0  & 2\\  
mono2 & 935  & 929 & 661 & 89 & 1 & 5 & 0 & 3\\ 
mono3 & 1,863 & 1,158 & 1,863 & 89 & 5 & 0 & 2 & 1\\ 
mono4 & 2,688 & 1,328 & 2,063 & 85 & 3 & 6 &2 & 4\\  
poly1 & 5,121 & 4,887 & 1,992 & 80 & 12 & 0 & 0 & 6\\  
poly2 & 1,450 & 1,426 & 449 & 75 & 16 & 2 & 0 & 5\\
poly3 & 11,687 & 6,611 & 3,165 & 58 & 35 & 0 & 1 & 5\\
poly4 & 40,298 & 9,400 & 3,754 & 61 & 23 & 5 & 1 & 9\\
Variant & 3,164 & 2,195 & 2,261 & 85 & 4 & 4 & 1 & 6\\  
Field   &510 & 379 & 421 & 78 & 9 & 2 & 2 & 9\\    
% \hline			
\end{tabular}
\caption{\label{manualverification} Results of class methods}
\label{manualverification:table:blah-blah}
\end{table*}

\end{itemize}

\subsection{Structural Methods}
In this set of methods the whole WN1.5 structure has been used to disambiguate. From HBil, all combinations of English words from 2 up to the maximum number of translations for each entry have been generated. The idea is to find as much common information between the corresponding EWs in WN1.5 as possible. On the extracted combinations, four experiments have been carried out resulting in the criteria described below:

\begin{itemize}
\item   Intersection criterion

Conditions:
	All EWs share at least one common synset in WordNet.
Link:
	SW is linked to all common synsets of its translations.
\item   Parent criterion

Conditions:
	A synset of an EW is direct parent of synsets corresponding to the rest of EWs.
Link:
	The SW is linked to all hyponym synsets\footnote{A previous experiment assigning SW only to the hypernym synset (assuming this would appropriately capture global information) resulted in  too general assignments.}
\item   Brother criterion 

Conditions:
	All EWs have synsets which are brothers respecting to a common parent.
Link:
	The SW is linked to all co-hyponym synsets.

\item   Distant hyperonymy criterion

Conditions:
	A synset of an EW is a distant hypernym of synsets of the rest of the EWs.
Link:
	The Spanish Word is linked to the lower-level (hyponym) synsets. 

\end{itemize}

As the results of all these criteria follow a structure like:

Spanish-Word $<$list-of-EW$>$ $<$list-of-synsets$>$, 

the  Structural Criteria have been subsequently pruned by deleting repeating entries subsumed by larger ones.

The overall results of Structural criteria are shown in table
\ref{structuralcriteria}.

\begin{table*}[htb]
\centering
\begin{tabular}{l|rrrrrrrr}
% \hline
\textsc{Criterion} & \textsc{\#Links} & \textsc{\#Synsets} & \textsc{\#Words} & \textsc{\%ok} & \textsc{\%ko} & \textsc{\%hypo} & \textsc{\%hyper} & \textsc{\%near}\\
\hline
inters  & 1,256 & 966  & 767 & 79 & 4 & 8  & 0 & 9\\
parent  & 1,432 & 1,210 & 788 & 51 & 3 & 30 & 0 & 14\\
brother & 2,202 & 1,645 & 672 & 57 & 5 & 22 & 0 & 16\\
distant & 1,846 & 1,522 & 866 & 60 & 4 & 23 & 0 & 13\\
% \hline			
\end{tabular}
\caption{\label{structuralcriteria} Overall results for the  Structural Criteria}
\label{structuralcriteria:table:blah-blah}
\end{table*}

A finer-grained experiment has been performed on the size of the
translation list. We have found that the larger this size is, the
higher is the precision obtained and, even more important, the lower
is the KO-ratio. The results for the case of intersection criterion
are shown in table \ref{intersectioncriteria}.

\begin{table}[htb]
\centering
\begin{tabular}{l|rrr}
% \hline
\textsc{\#Words} & \textsc{\%ok} & \textsc{\%ko} & \textsc{\%hypo}\\
\hline
2 & 81,39 & 3,48 & 1,51\\
3 & 91,89 & 0,0  & 5,4\\
4 & 94,4  & 0,0  & 0,0\\
% \hline			
\end{tabular}
\caption{\label{intersectioncriteria} Results for the Intersection Criteria}
\label{intersectioncriteria:table:blah-blah}
\end{table}

\subsection{Conceptual Distance Methods}

Taking as reference a structured hierarchical net, conceptual distance tries to provide a basis for determining closeness in meaning among words. Conceptual distance between two concepts is defined in \cite{Rada+'89} as the length of the shortest path that connects the concepts in a hierarchical semantic net. In a similar approach, \cite{Sussna'93} employs the notion of conceptual distance between network nodes in order to improve precision during document indexing. Following these ideas, \cite{Agirre+'94} describe a new conceptual distance formula for automatic spelling correction and \cite{Rigau'95}, using this conceptual distance formula, presents a methodology to enrich dictionary senses with semantic tags extracted from WordNet. The same measure is used in \cite{Rigau+'95} for linking taxonomies extracted from DGILE and LDOCE and in \cite{Rigau+'97} as one of the methods for the Genus Sense Disambiguation problem in DGILE. Conceptual density, a more complex semantic measure among words is defined in \cite{Agirre+Rigau'95} and used in \cite{Agirre+Rigau'96} as a proposal for WSD of the Brown Corpus.
The Conceptual Distance formula used in this work, also described in
\cite{Agirre+'94} is shown in Figure 5. 

\begin{equation}
\renewcommand\arraystretch{0.7}
dist(w_{1},w_{2}) = \min_{\begin{array}{c}  
                             \scriptstyle c_{1_{i}} \in w_{1} \\
                             \scriptstyle c_{2_{j}} \in w_{2}
			  \end{array}}
		        \sum_{\begin{array}{c}  
                             		\scriptstyle c_{k} \in \\
					\scriptstyle path(c_{1_{i}},
							      c_{2_{j}})
			       \end{array}}
                            \frac{1}{depth(c_{k})} 
\end{equation}
\centerline{Figure 5: Conceptual distance formula}\\
where Wi are words and Ci are synsets representing those words. Conceptual Distance between two words depends on the length of the shortest path that connects the concepts and the specificity of the concepts in the path. Then, providing two words, the application of the Conceptual Distance formula selects those closer concepts which represent them. 

Following this approach, three different methods have been applied.

\subsubsection{Using Co-occurrence words collected from DGILE (CD1)}

Following \cite{Wilks+'93} two words are coocurrent in a dictionary if they appear in the same definition. For DGILE, a lexicon of 300,062 coocurrence pairs among 40,193 Spanish word forms was derived and the afinity between these pairs was measured by means of the Association Ratio (AR), which can be used as a fine grained CS. 

Then, the Conceptual Distance formula for all those pairs has been computed using HBil and the nominal part of WN. 

\subsubsection{Using Headword and genus of DGILE (CD2)}
Computing the Conceptual Distance formula on the headword and the genus term of  92,741 nominal definitions of DGILE dictionary (only 32,208 with translation to English).

\subsubsection{Using Spanish entries with multiple translations in the bilingual dictionary (CD3)}
In this case, we have derived a small but closely related lexicon of 3,117 translation equivalents with multiple translations from the Spanish/English direction of the bilingual dictionary (only 2,542 with connection to WordNet1.5).

Table \ref{conceptualdistance} summarizes the performance of the three Conceptual Distance methods.
 
\begin{table*}[htb]
\centering
\begin{tabular}{l|rrrrrrrr}
% \hline
\textsc{Criterion} & \textsc{\#Links} & \textsc{\#Synsets} & \textsc{\#Words} & \textsc{\%ok} & \textsc{\%ko} & \textsc{\%hypo} & \textsc{\%hyper} & \textsc{\%near}\\
\hline
CD - 1 & 23,828 & 11,269 & 7,283 & 56 & 38 & 3 & 2 & 2\\
CD - 2 & 24,739 & 12,709 & 10,300 & 61 & 35 & 0 & 0 & 3\\
CD - 3 & 4,567 & 3,089 & 2,313 & 75 & 12 & 0 & 2 & 8\\
% \hline			
\end{tabular}
\caption{\label{conceptualdistance} Performace of Conceptual Distance methods}
\label{conceptualdistance:table:blah-blah}
\end{table*}

\section{Combining methods}

Collecting those synsets produced by the methods described above with an accuracy greater than 85\% (mono1, mono2, mono3, mono4, variants, field) we obtain a preliminary version of the Spanish WordNet containing 10,982 connections (1,777 polysemous) among 7,131 synsets and 8,396 Spanish nouns with an overall CS of 87,4\%. However, combining the discarded methods we can take profit of portions of them precise enough to be acceptable.

All files resulting from discarded methods were crossed and their
intersections were calculated. Using VI, a manual inspection of
samples from each intersection was carried out. Results are shown in
table \ref{sampleresults}.

\begin{table*}[htb]
\centering
\begin{tabular}{l|rrrrrrrrrr}
% \hline
\textsc{} & method2 \\
\hline
method1 &  & cd1 & cd2 & cd3 & dist & fath & p1 & p2 & p3 & p4\\
bro &size    & 855 & 828   & 435        & 449  	& 405	& {\bf 76} 	& {\bf 107} 	& 0 		& 1,872\\
    &\%ok    & 70  & 71    & 79         & 58   	& 6	& {\bf 86} 	& {\bf 89} 	& 0 		& 67\\
cd1 & size   & 0   & 15,736 & {\bf 1,849} & 576  	& 419	& {\bf 2,076} 	& {\bf 556}	& 3,146 		& 15,105\\
    &\%ok    & 0   & 79    & {\bf 85}   & 68   	& 71	& {\bf 86}	& {\bf 86} 	& 72 		& 64\\
cd2 & size   & 0   & 0 	   & {\bf 2,401} & 571  	& 428	& {\bf 2,536} 	& {\bf 592} 	& 3,777 		& 13,246\\
    & \%ok   & 0   & 0     & {\bf 86}   & 71   	& 72	& {\bf 88} 	& {\bf 86} 	& 75 		& 67\\
cd3 & size   & 0   & 0 	   & 0 		& 391  	& 325	& {\bf 205} 	& {\bf 180} 	& {\bf 215} 	& 3,114\\ 
    & \%ok   & 0   & 0 	   & 0 		& 79   	& 80   	& {\bf 95}	& {\bf 95}   	& {\bf 100} 	& 77\\
dist & size  & 0   & 0     & 0 		& 0    	& 1,432 	& 69 	   	& 68	   	& 0 		& 1,463\\
     & \%ok  & 0   & 0     & 0 		& 0    	& 67    & 78 		& 7    		& 0 		& 65\\
fath & size  & 0   & 0     & 0 		& 0    	& 0     & 69 		& 61 		& 0 		& 1,101\\
     & \%ok  & 0   & 0     & 0 		& 0    	& 0     & 77 		& 70 		& 0 		& 67\\
p1 & size    & 0   & 0     & 0 		& 0    	& 0     & 0		& 0 		& {\bf 77} 	& {\bf 178}\\
   & \%ok    & 0   & 0     & 0 		& 0	& 0	& 0		& 0 		& {\bf 100} 	& {\bf 88}\\
p2 & size    & 0   & 0     & 0 		& 0 	& 0	& 0		& 0 		& 28 		& {\bf 78}\\
   & \%ok    & 0   & 0     & 0 		& 0 	& 0	& 0		& 0 		& 77 		& {\bf 96}\\ 
% \hline			
\end{tabular}
\caption{\label{sampleresults} Results combining methods}
\label{sampleresults:table:blah-blah}
\end{table*}

In bold appear intersections with a CS greater than 85\%. Up to 7,244 connections (2,075 polysemous) can be selected with  85.63\% CS, 4,553 of which are new with an overall CS of 84\%  resulting in a 41\% increase. It must be pointed out that 1,308 new connections are polysemous.

Then a second version of the Spanish WordNet has been obtained containing 15,535 connections (3,373 polysemous) among 10,786 synsets and 9,986 Spanish nouns with a final accuracy of 86,4\%. Table \ref{overallfigures} shows the overall figures of the resulting SpWNs.

\begin{table*}[htb]
\centering
\begin{tabular}{l|rrrrr}
% \hline
\textsc{Criterion} & \textsc{\#Links} & \textsc{\#Synsets} & \textsc{\#Words} & \textsc{\#CS} & \textsc{\#Poly Links}\\
\hline
SpWN v.0.0 & 10,982 &  7,131 & 8,396 & 87.4 & 1,777 \\
Combination & 7,244 & 5,852 & 3,939 & 85.6 & 2,075 \\
SpWN v.0.1 & 15,535 & 10,786 & 9,986 & 86.4 & 3,373 \\
% \hline			
\end{tabular}
\caption{\label{overallfigures} Overall Figures of SpWNs}
\label{overallfigures:table}
\end{table*}

\section{Conclusions}
An approach for building multilingual Wordnets combining a variety of lexical sources as well as a variety of methods has been proposed which tries to take profit of the existing WN1.5 for attaching words from other languages in a way guided mainly by the content of bilingual lexical sources. 

A central issue of our approach is the combination of methods and sources in a way that the accuracy of the data obtained from the combined methods overcomes the accuracy obtained from the individual ones.  Several families of methods have been tested, each of them bearing its own CS. Only those methods offering a result over a threshold (85\%) have been considered.

In a second phase of our experiments, intersections between the results provided by the different individual methods have been performed. It is clear that  valuable sets of entries, owning an insufficient, in some cases rather bad, individual CS, can be, however, extracted if they occur as a combination of several methods. In this way, using the same threshold, the amount of synsets attached to Spanish entries has increased. It must be pointed out that some of these new connections correspond to highly polysemous words.

The approach seems to be extremely promising, attaching up to 75\% of
reachable Spanish nouns and 55\% of reachable WN1.5 synsets. Currently we are performing complementary experiments for extending the approach for covering other lexical sources, specially wider-coverage bilinguals. 

Other lines of research we are following by now include: 1) dealing with mismatches, i.e.,  when coming from different method/source an Spanish word is assigned to different synsets. If in the former case the overall CS increases,  in the last one it should decrease. 2) A fine grained cross-comparison of methods and sources (intersections of more than two classes, decomposition of classes into finer ones, etc.) will be performed to obtain a more precise classification and CS assignment. 3) We are trying to obtain an empirical method for CS calculation of intersections. Methods based on bayesian inference networks or quasiprobabilistic approaches has been tested giving promising results.

\bibliographystyle{ranlp}
\bibliography{bib_data}

\end{document}